\documentclass[11pt,letterpaper]{article}
\usepackage{acl2010}
\usepackage{times}
\usepackage{latexsym}

\usepackage{helvet}
\usepackage{courier}
\usepackage{times}
\usepackage{url}
\usepackage{latexsym}
\usepackage{multirow}
\usepackage{algorithm}
\usepackage{algorithmic}
\usepackage{graphics}
\usepackage{graphicx}
\usepackage{epsfig}
\usepackage{amssymb}
\usepackage{amsmath}
\usepackage{tablefootnote}
\usepackage{listings}

\usepackage{helvet}
\usepackage{courier}
\usepackage{times}
\usepackage{url}
\usepackage{latexsym}
\usepackage{multirow}
\usepackage{algorithm}
\usepackage{algorithmic}
\usepackage{graphics}
\usepackage{graphicx}
\usepackage{epsfig}
\usepackage{amssymb}
\usepackage{amsmath}
\usepackage{tablefootnote}

\DeclareMathOperator*{\argmax}{\arg\!\max}

\makeatletter
\newcommand*{\rom}[1]{\expandafter\@slowromancap\romannumeral #1@}
\makeatother

\def\ignore#1{}

\def\ignore#1{}
\newcommand\floor[1]{\lfloor#1\rfloor}

\graphicspath{{figures/}}

\title{Enhancing Sentence Relation Modeling with Auxiliary Character-level Embedding}

\author{Peng Li \\
  Computer Science and Engineering \\
  University of Texas at Arlington \\
  {\tt jerryli1981@gmail.com} \\\And
  Heng Huang \\
  Computer Science and Engineering \\
  University of Texas at Arlington \\
  {\tt heng@uta.edu} \\}

\begin{document}

\maketitle
\begin{abstract}
Neural network based approaches for sentence relation modeling automatically generate hidden matching features from raw sentence pairs. However, the quality of matching feature representation may not be satisfied due to complex semantic relations such as entailment or contradiction. To address this challenge,  we propose a new deep neural network architecture that jointly leverage pre-trained word embedding and auxiliary character embedding to learn sentence meanings.  The two kinds of word sequence representations as inputs into multi-layer bidirectional LSTM to learn enhanced sentence representation. After that, we construct matching features followed by another temporal CNN to learn high-level hidden matching feature representations.  Experimental results demonstrate that our approach consistently outperforms the existing methods on standard evaluation datasets.
\end{abstract}

\section{Introduction}
Traditional approaches~\cite{Lai2014,Zhao2014ecnu,Jimenez2014} for sentence relation modeling tasks such as paraphrase identification, question answering, recognized textual entailment and semantic textual similarity prediction usually build the supervised model using a variety of hand crafted features. Hundreds of features generated at different linguistic levels are exploited to boost classification. With the success of deep learning,  there has been much interest in applying deep neural network based techniques to further improve the prediction performances ~\cite{Socher2011,Mohit2014,Yin2015Multi}. 

A key component of deep neural network is word embedding which serve as an lookup table to get word representations. From low level NLP tasks such as language modeling, POS tagging, name entity recognition, and semantic role labeling~\cite{Collobert2011,Mikolov2013}, to high level tasks such as machine translation, information retrieval and semantic analysis~\cite{kalchbrenner2013,Socher2011dynamic,Tai2015}. Deep word representation learning has demonstrated its importance for these tasks.  All the tasks get performance improvement via further learning either word level representations or sentence level representations. On the other hand, some researchers have found character-level convolutional networks~\cite{kim2015character,zhang2015character} are useful in extracting information from raw signals for the task such as language modeling or text classification. 

\begin{figure}[!htb]
\centering
\includegraphics[width=8cm,height=7cm]{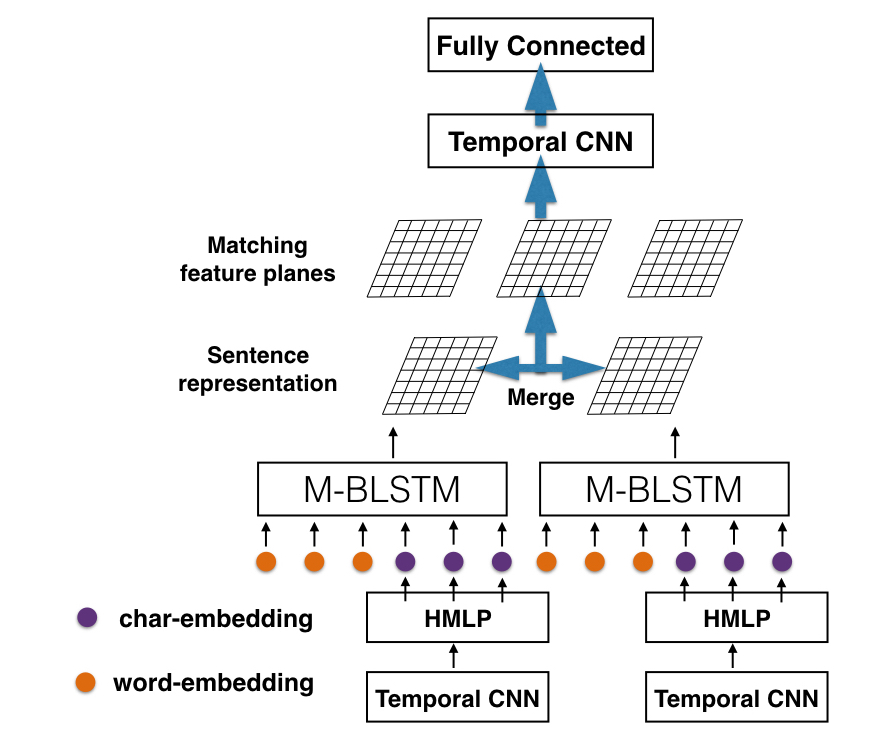}
\caption{Neural Network Architecture for Deep Matching Feature Learning. M-BLSTM is Multi-layer Bidirectional LSTM. Orange color represents sequence representations that concatenating pre-trained word vectors. Purple color represents sequence representation concatenating word vectors that generating from character-level convolutional network and HMLP.}
\label{fig:arc}
\end{figure}

In this work, we focus on deep neural network based sentence relation modeling tasks. We explore treating each sentence as a kind of raw signal at character level, and applying temporal (one-dimensional) Convolution Neural Network (CNN)~\cite{Collobert2011}, Highway Multilayer Perceptron (HMLP) and multi-layer bidirectional LSTM (Long Short Term Memory)~\cite{Alex2013} to learn sentence representations. We propose a new deep neural network architecture that jointly leverage pre-trained word embedding and character embedding to represent the meaning sentences. More specifically, our new approach first generates two kinds of word sequence representations. One kind of sequence representations are the composition of pre-trained word vectors. The other kind of sequence representation comprise word vectors that generating from character-level convolutional network. We then inject the two sequence representations into bidirectional LSTM, which means forward directional LSTM accept pre-trained word embedding output and backward directional LSTM accept auxiliary character CNN embedding output. The final sentence representation is the concatenation of the two direction. After that, we construct matching features followed by another temporal CNN to learn high-level hidden matching feature representations. Figure~\ref{fig:arc} shows the neural network architecture for general sentence relation modeling. 

Our model shows that when trained on small size datasets, combining pre-trained word embeddings with auxiliary character-level embedding can improve the sentence representation. Word embeddings can help capturing general word semantic meanings, whereas char-level embedding can help modeling task specific word meanings. Note that auxiliary character-level embedding based sentence representation do not require the knowledge of words or even syntactic structure of a language. The enhanced sentence representation generated by multi-layer bidirectional LSTM will encapsulate the character and word levels informations. Furthermore, it may enhance matching features that generated by computing similarity measures on sentence pairs. Quantitative evaluations on standard dataset demonstrate the effectiveness and advantages of our method.

\section{Character-level Convolutional Neural Network}
Besides pre-trained word vectors, we are also interested in generating word vectors from characters. To achieve that, we leverage deep convolutional neural network(ConvNets).  The model accepts a sequence of encoded characters as input. The encoding si done by prescribing an alphabet of size $m$ for the input language, and then quantize each character using one-hot encoding. Then, the sequence of characters is transformed to a sequence of such $m$ sized vectors with fixed length $l_0$. Any character exceeding length $l_0$ is ignored, and any characters that are not in the alphabet are quantized as all-zero vectors. The alphabet used in our model consists of 36 characters, including 26 english letters and 10 digits. Below, we will introduce character-level temporal convolution neural network. 

\subsection{Temporal Convolution}
Temporal Convolution applies one-dimensional convolution over an input sequence. The one-dimensional convolution is an operation between a vector of weights $\bold{m} \in  \mathbb{R}^{m}$ and a vector of inputs viewed as a sequence $\bold{x} \in  \mathbb{R}^{n}$. The vector $\bold{m}$ is the \emph{filter} of the convolution. Concretely, we think of $\bold{x}$ as the input token and $\bold{x}_i \in \mathbb{R}$ as a single feature value associated with the $i$-th character in this token. The idea behind the one-dimensional convolution is to take the dot product of the vector $\bold{m}$ with each $m$-gram in the token $\bold{x}$ to obtain another sequence $\bold{c}$:
\begin{equation}
\bold{c}_j = \bold{m}^{T}\bold{x}_{j-m+1:j}\,.
\end{equation}

Usually, $\bold{x}_i$ is not a single value, but a $d$-dimensional vector so that $\bold{x} \in \mathbb{R}^{d \times n}$. There exist two types of 1d convolution operations. One is called Time Delay Neural Networks (TDNNs). The other one was introduced by~\cite{Collobert2011}. In TDNN, weights $\bold{m} \in \mathbb{R}^{d \times m}$ form a matrix. Each row of $\bold{m}$ is convolved with the corresponding row of $\bold{x}$. In~\cite{Collobert2011} architecture, a sequence of length $n$ is represented as:
\begin{equation}
\bold{x}_{1:n} = \bold{x}_1 \oplus \bold{x}_2 \cdots \oplus \bold{x}_n\,,
\end{equation}
where $\oplus$ is the concatenation operation. In general, let $\bold{x}_{i:i+j}$ refer to the concatenation of characters $\bold{x}_i, \bold{x}_{i+1}, \dots, \bold{x}_{i+j}$. A convolution operation involves a filter $\bold{w} \in \mathbb{R}^{hk}$, which is applied to a window of $h$ characters to produce the new feature. For example, a feature $c_i$ is generated from a window of characters $\bold{x}_{i:i+h-1}$ by:
\begin{equation}
c_i = f( \bold{w} \cdot \bold{x}_{i:i+h-1} + b )\,.
\end{equation}
Here $b \in \mathbb{R}$ is a bias term and $f$ is a non-linear function such as the thresholding function $f(x) = max\{0,x\}$. This filter is applied to each possible window of characters in the sequence $\{ \bold{x}_{1:h},  \bold{x}_{2:h+1}, \dots, \bold{x}_{n-h+1:n} \}$ to produce a feature map:
\begin{equation}
\bold{c} = [ c_1, c_2, \dots, c_{n-h+1}]\,,
\end{equation}
with $\bold{c} \in \mathbb{R}^{n-h+1}$.

\subsection{Highway MLP}
On top of convolutional neural network layers, we build another Highway Multilayer Perceptron (HMLP) layer to further enhance character-level word embeddings. Conventional MLP applies an affine transformation followed by a nonlinearity to obtain a new set of features:
\begin{equation}
\bold{z} = g(\bold{W}\bold{y} + \bold{b})\,.
\end{equation}
One layer of a highway network does the following:
\begin{equation}
\bold{z} = \bold{t} \odot g(\bold{W}_H\bold{y} + \bold{b}_H) + (1-\bold{t} ) \odot \bold{y}\,,
\end{equation}
where $g$ is a nonlinearity, $\bold{t} = \sigma(\bold{W}_T\bold{y} + \bold{b}_T)$ is called as the transform gate, and $(1-\bold{t})$ is called as the carry gate. Similar to the memory cells in LSTM networks, highway layers allow adaptively carrying some dimensions of the input directly to the input for training deep networks.

\section{Multi-Layer Bidirectional LSTM}
Now that we have two kinds of word sequence representations. One kind of sequence representations are the composition of pre-trained word vectors. The other kind of sequence representation comprise word vectors that generating from character-level convolutional network. We can inject the two sequence representations into bidirectional LSTM to learn sentence representation. More specifically, forward directional LSTM accept pre-trained word embedding output and backward directional LSTM accept character CNN embedding output. The final sentence representation is the concatenation of the two direction. 

\subsection{RNN vs LSTM}
Recurrent neural networks (RNNs) are capable of modeling sequences of varying lengths via the recursive application of a transition function on a hidden state. For example, at each time step $t$, an RNN takes the input vector $\bold{x}_t \in \mathbb{R}^n$ and the hidden state vector $\bold{h}_{t-1} \in \mathbb{R}^m$, then applies affine transformation followed by an element-wise nonlinearity such as hyperbolic tangent function to produce the next hidden state vector $\bold{h}_t$:
\begin{equation}
\bold{h}_t = \tanh (\bold{W}\bold{x}_t + \bold{U}\bold{h}_{t-1} + \bold{b} ) \,.
\end{equation}

A major issue of RNNs using these transition functions is that it is difficult to learn long-range dependencies during training step because the components of the gradient vector can grow or decay exponentially~\cite{Bengio1994}.

The LSTM architecture~\cite{Sepp1998} addresses the problem of learning long range dependencies by introducing a memory cell that is able to preserve state over long periods of time. Concretely, at each time step $t$, the LSTM \emph{unit} can be defined as a collection of vectors in $\mathbb{R}^d$: an \emph{input gate} $\bold{i}_t$, a \emph{forget gate} $\bold{f}_t$, an \emph{output gate} $\bold{o}_t$, a \emph{memory cell} $\bold{c}_t$ and a hidden state $\bold{h}_t$.  We refer to $d$ as the \emph{memory dimensionality} of the LSTM. One step of an LSTM takes as input $\bold{x}_t$, $\bold{h}_{t-1}$, $\bold{c}_{t-1}$ and produces $\bold{h}_t$, $\bold{c}_t$ via the following transition equations:
\begin{equation}
\begin{split}
\bold{i}_t &= \sigma (\bold{W^{(i)}}\bold{x}_t + \bold{U^{(i)}}\bold{h}_{t-1} + \bold{b^{(i)}} ) \,,\\
\bold{f}_t &= \sigma (\bold{W^{(f)}}\bold{x}_t + \bold{U^{(f)}}\bold{h}_{t-1} + \bold{b^{(f)}} ) \,,\\
\bold{o}_t &= \sigma (\bold{W^{(o)}}\bold{x}_t + \bold{U^{(o)}}\bold{h}_{t-1} + \bold{b^{(o)}} ) \,,\\
\bold{u}_t &= \tanh (\bold{W^{(u)}}\bold{x}_t + \bold{U^{(u)}}\bold{h}_{t-1} + \bold{b^{(u)}} ) \,,\\
\bold{c}_t &= \bold{i}_t \odot \bold{u}_t + \bold{f}_t \odot \bold{c}_{t-1}\,,\\
\bold{h}_t &= \bold{o}_t \odot \tanh(\bold{c}_t) \,,\\
\end{split}
\end{equation}
where $\sigma(\cdot)$ and $\tanh(\cdot)$ are the element-wise sigmoid and hyperbolic tangent functions, $\odot$ is the element-wise multiplication operator.

\subsection{Model Description}
One shortcoming of conventional RNNs is that they are only able to make use of previous context. In text entailment, the decision is made after the whole sentence pair is digested. Therefore, exploring future context would be better for sequence meaning representation.  Bidirectional RNNs architecture~\cite{Alex2013} proposed a solution of making prediction based on future words. At each time step $t$, the model maintains two hidden states, one for the left-to-right propagation $\overrightarrow{\bold{h}_t}$ and the other for the right-to-left propagation $\overleftarrow{\bold{h}_t}$. The hidden state of the Bidirectional LSTM is the concatenation of the forward and backward hidden states. The following equations illustrate the main ideas:
\begin{equation}
\begin{split}
\overrightarrow{\bold{h}_t} &= \tanh (\overrightarrow{\bold{W}}\bold{x}_t + \overrightarrow{\bold{U}}\overrightarrow{\bold{h}}_{t-1} + \overrightarrow{\bold{b}} ) \\
\overleftarrow{\bold{h}_t} &= \tanh (\overleftarrow{\bold{W}}\bold{x}_t + \overleftarrow{\bold{U}}
\overleftarrow{\bold{h}}_{t+1} + \overleftarrow{\bold{b}} ) \,.\\
%\bold{h}_t &= \bold{W}_l\overrightarrow{\bold{h}_t} +  \bold{W}_r\overleftarrow{\bold{h}_t} + \bold{b}
\end{split}
\end{equation}

Deep RNNs can be created by stacking multiple RNN hidden layer on top of each other, with the output sequence of one layer forming the input sequence for the next. Assuming the same hidden layer function is used for all $N$ layers in the stack, the hidden vectors $\bold{h}^n$ are iteratively computed from $n=1$ to $N$ and $t=1$ to $T$:
\begin{equation}
\bold{h}^n_t = \tanh (\bold{W} \bold{h}^{n-1}_t + \bold{U} \bold{h}^{n}_{t-1} +  \bold{b} )\,.\\
\end{equation}

Multilayer bidirectional RNNs can be implemented by replacing each hidden vector $\bold{h}^n$ with the forward and backward vectors $\overrightarrow{\bold{h}^n}$ and $\overleftarrow{\bold{h}^n}$, and ensuring that every hidden layer receives input from both the forward and backward layers at the level below. Furthermore, we can apply LSTM memory cell to hidden layers to construct multilayer bidirectional LSTM.

Finally, we can concatenate sequence hidden matrix $\overrightarrow{\bold{M}} \in \mathbb{R}^{n \times d}$ and reversed sequence hidden matrix $\overleftarrow{\bold{M}} \in \mathbb{R}^{n \times d}$ to form the sentence representation. We refer to $n$ is the number of layers, $d$ as the \emph{memory dimensionality} of the LSTM. In the next section, we will use the two matrixs to generate matching feature planes via linear algebra operations.

\section{Learning from Matching Features}
Inspired by~\cite{Tai2015}, we apply element-wise merge to first sentence matrix $M_1 \in \mathbb{R}^{n \times 2d}$ and second sentence matrix $M_2 \in \mathbb{R}^{n \times 2d}$. Similar to previous method, we can define two simple matching feature planes (\emph{FPs}) with below equations:
\begin{equation}
\begin{split}
FP_1 &= M_1  \odot M_2 \,,\\
FP_2 &= | M_1 - M_2 | \,,\\
\end{split}
\end{equation}
where $\odot$ is the element-wise multiplication. The $FP_1$ measure can be interpreted as an element-wise comparison of the signs of the input representations. The $FP_2$ measure can be interpreted as the distance between the input representations.

In addition to the above measures, we also found the following feature plane can improve the performance:
\begin{equation}
\begin{split}
FP_3 &= 1dConv(Reshape(Join(M_1, M_2))) \,,\\
\end{split}
\end{equation}
In $FP_3$, the $1dConv$ means one-dimensional convolution. Join mean concatenate the two representation. The intuition behind $FP_3$ is let the one-dimensional convolution preserves the common information between sentence pairs. 

\subsection{Reshape Feature Planes}
Recall that the multi-layer bidirectional LSTM generates sentence representation matrix $\bold{M} \in \mathbb{R}^{n \times 2d}$ by concatenating sentence hidden matrix $\overrightarrow{\bold{M}} \in \mathbb{R}^{n \times d}$ and reversed sentence hidden matrix $\overleftarrow{\bold{M}} \in \mathbb{R}^{n \times d}$.  Then we conduct element-wise merge to form feature plane $\bold{M}_{fp} \in \mathbb{R}^{n \times 2d}$. Therefore, the final input into temporal convolution layer is a 3D tensor $\bold{I} \in \mathbb{R}^{f \times n \times 2d}$, where $f$ is the number of matching feature plane, $n$ is the number of layers, $d$ as the \emph{memory dimensionality} of the LSTM. Note that the 3D tensor convolutional layer input $\bold{I}$ can be viewed as an image where each feature plane is a channel. In computer vision and image processing communities, the spatial 2D convolution is often used over an input image composed of several input planes. In experiment section, we will compare 2D convolution with 1D convolution. In order to facilitate temporal convolution, we need reshape $\bold{I}$ to 2D tensor.

\subsection{CNN Topology}
The matching feature planes can be viewed as channels of images in image processing. In our scenario, these feature planes hold the matching information. We will use temporal convolutional neural network to learn hidden matching features. The mechanism of temporal CNN here is the same as character-level temporal CNN. However, the kernels are totally different. 

\begin{figure*}[!htb]
\centering
\begin{minipage}{.46\textwidth}
\includegraphics[width=8.6cm,height=7.4cm]{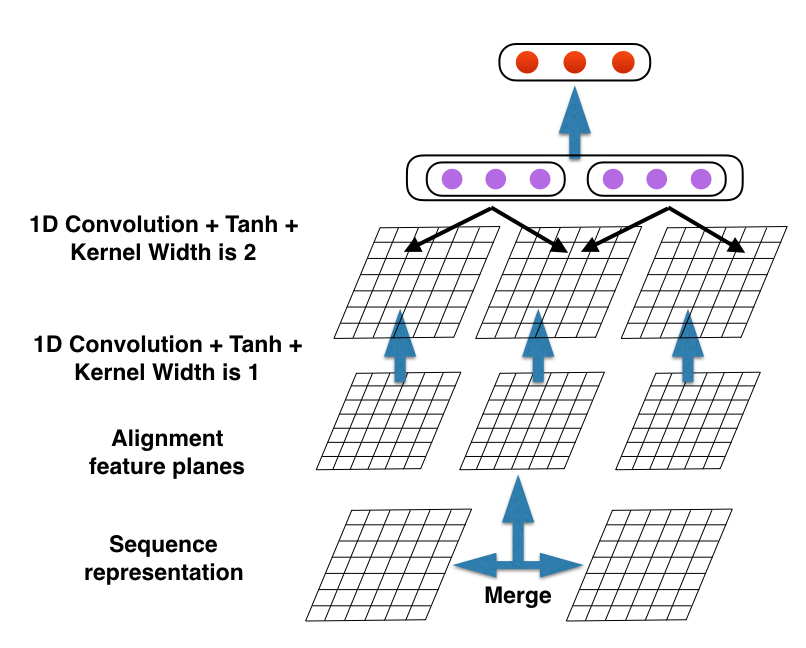}
\caption{CNN Topology \rom{1}}
\label{fig:cnnTopo1}
\end{minipage}%
\qquad
\begin{minipage}{.46\textwidth}
\includegraphics[width=8.6cm,height=7.4cm]{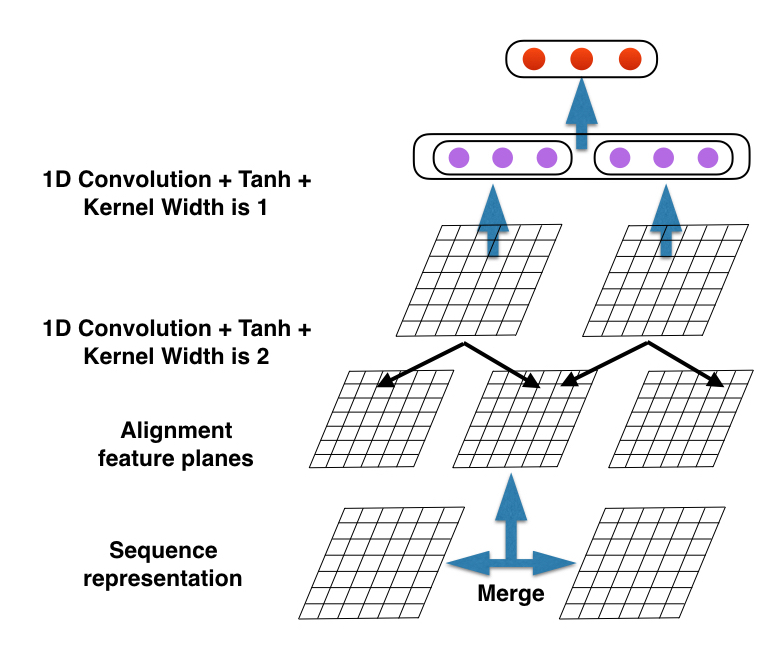}
\caption{CNN Topology \rom{2}}
\label{fig:cnnTopo2}
\end{minipage}
\end{figure*}

It's quite important to design a good topology for CNN to learn hidden features from heterogeneous feature planes. After several experiments, we found two topological graphs can be deployed in the architecture. Figure~\ref{fig:cnnTopo1} and Figure~\ref{fig:cnnTopo2} show the two CNN graphs. In Topology \rom{1}, we stack temporal convolution with kernel width as 1 and tanh activation on top of each feature plane. After that, we deploy another temporal convolution and tanh activation operation with kernel width as 2. In Topology \rom{2}, however, we first stack temporal convolution and tanh activation with kernel width as 2. Then we deploy another temporal convolution and tanh activation operation with kernel width as 1. Experiment results demonstrate that the Topology \rom{1} is slightly better than the Topology \rom{2}. This conclusion is reasonable. The feature planes are heterogeneous. After conducting convolution and tanh activation transformation, it makes sense to compare values across different feature planes.

\section{Experiments}
We selected two related sentence relation modeling tasks: semantic relatedness task, which measures the degree of semantic relatedness of a sentence pair by assigning a relatedness score ranging from 1 (completely unrelated) to 5 ( very related); and textual entailment task, which determines whether the truth of a text entails the truth of another text called hypothesis. We use standard SICK (Sentences Involving Compositional Knowledge) dataset~\footnote{\url{http://alt.qcri.org/semeval2014/task1/index.php?id=data-and-tools}}  for evaluation. It consists of about 10,000 English sentence pairs annotated for relatedness in meaning and entailment.

\subsection{Hyperparameters and Training Details}
We first initialize our word representations using publicly available 300-dimensional Glove word vectors~\footnote{\url{http://nlp.stanford.edu/projects/glove/}}. LSTM memory dimension is 100, the number of layers is 2. On the other hand, for CharCNN model we use threshold activation function on top of each temporal convolution and max pooling pairs . The CharCNN input frame size equals alphabet size, output frame size is 100. The maximum sentence length is 37. The kernel width of each temporal convolution is set to 3, the step is 1, the hidden units of HighwayMLP is 50. Training is done through stochastic gradient descent over shuffled mini-batches with the AdaGrad update rule~\cite{Duchi2011}. The learning rate is set to 0.05. The mini-batch size is 25. The model parameters were regularized with a per-minibatch L2 regularization strength of $10^{-4}$.  Note that word embeddings were fixed during training.

\subsection{Objective Functions}
The task of semantic relatedness prediction tries to measure the degree of semantic relatedness of a sentence pair by assigning a relatedness score ranging from 1 (completely unrelated) to 5 (very related).  More formally, given a sentence pair, we wish to predict a real-valued similarity score in a range of $[1, K ]$, where $K > 1$ is an integer. The sequence ${1, 2, . . . , K}$ is the ordinal scale of similarity, where higher scores indicate greater degrees of similarity. We can predict the similarity score $\hat{y}$ by predicting the probability that the learned hidden representation $x_h$ belongs to the ordinal scale. This is done by projecting an input representation onto a set of hyperplanes, each of which corresponds to a class. The distance from the input to a hyperplane reflects the probability that the input will located in corresponding scale.

Mathematically, the similarity score $\hat{y}$ can be written as:
\begin{eqnarray}
\hat{y} &=& r^T \cdot \hat{p}_\theta (y |x_h) \nonumber\\
&=&  r^T \cdot softmax(W \cdot x_h + b) \nonumber\\
                                    &=& r^T \cdot  \frac{e^{W_i x_h +b_i}}{\sum_j e^{W_j x_h +b_j}} 
\end{eqnarray}
where $r^T = [1~2 \dots K]$ and the weight matrix $W$ and $b$ are parameters.

In order to introduce the task objective function, we define a sparse target distribution $p$ that satisfies $y = r^Tp$:
\begin{equation}
p_i =
\left\{
	\begin{array}{ll}
		y-\floor{y},  &  i = \floor{y}+1 \\
		\floor{y}-y+1, & i = \floor{y} \\
		0 & otherwise
	\end{array}
\right.
\end{equation}
where $1 \leq i \leq K$. The objective function then can be defined as the regularized KL-divergence between $p$ and $p_{\theta}$:
\begin{equation}
J(\theta) = -\frac{1}{m} \sum_{k=1}^m KL(p^{(k)} || p_{\theta}^{{k}}) + \frac{\lambda}{2} ||\theta ||_2^2\,,
\label{simcost}
\end{equation}
where $m$ is the number of training pairs and the superscript $k$ indicates the $k$-th sentence pair~\cite{Tai2015}.

Referring to textual entailment recognition task,  we want to maximize the likelihood of the correct class. This is equivalent to minimizing the negative log-likelihood (NLL). More specifically, the label $\hat{y}$ given the inputs $x_h$ is predicted by a softmax classifier that takes the hidden state $h_j$ at the node as input:
\begin{equation}
\begin{split}
\hat{p}_\theta (y | x_h) &= softmax(W \cdot x_h + b) \\
\hat{y} &= \argmax_y \hat{p}_\theta (y | x_h)
\end{split}
\end{equation}

After that, the objective function is the negative log-likelihood of the true class labels $y^k$:
\begin{equation}
J(\theta) = -\frac{1}{m} \sum_{k=1}^m \log\hat{p}_\theta (y^k | x_h^k) + \frac{\lambda}{2} ||\theta ||_2^2\,,
\label{entcost}
\end{equation}
where $m$ is the number of training pairs and the superscript $k$ indicates the $k$th sentence pair.

\subsection{Results and Discussions}
Table~\ref{tab:pear} and~\ref{tab:acc} show the Pearson correlation and accuracy comparison results of semantic relatedness and text entailment tasks. We can see that combining CharCNN with multi-layer bidirectional LSTM yields better performance compared with other traditional machine learning methods such as SVM and MaxEnt approach~\cite{Thomas2014,Lai2014} that served with many handcraft features. Note that our method doesn't need extra handcrafted feature extraction procedure. Also our method doesn't leverage external linguistic resources such as wordnet or parsing which get best results in~\cite{Tai2015}. More importantly, both task prediction results close to the state-of-the-art results. It proved that our approaches successfully simultaneously predict \emph{heterogeneous} tasks. Note that for semantic relatedness task, the latest research~\cite{Tai2015} proposed a tree-structure based LSTM, the Pearson correlation score of their system can reach 0.863. Compared with their approach, our method didn't use dependency parsing and can be used to predict tasks contains multiple languages. 

We hope to point out that we implemented the method in \cite{Tai2015}, but the results are not as good as our method. Here we use the results reported in their paper. Based on our experiments, we believe the method in \cite{Tai2015} is very sensitive to the initializations, thus it may not achieve the good performance in different settings. However, our method is pretty stable which may benefit from the joint tasks training.

\ignore{
\begin{figure*}[!htb]
\centering
\includegraphics[width=11cm,height=6.2cm]{curve.png}
\caption{Accuracy and Pearson Correlation results versus the number of training epochs on develop dataset.}
\label{fig:curve}
\end{figure*}}

\begin{table*}[!htb]
\begin{center}
\begin{tabular}{c|p{1cm}|p{1cm}|p{5cm}}
\hline
Method & \multicolumn{1}{c|}{Pearson Correlation} & \multicolumn{1}{c|}{Features} & \multicolumn{1}{c}{Reported in}\\
\hline
\multirow{1}{*}{MaxEnt}  & 0.799 & 137 & \cite{Lai2014} \\
\cline{2-3}
\hline
\multirow{1}{*}{Decision tree}  & 0.804 & 214 & \cite{Jimenez2014}\\
\cline{2-3}
\hline
\multirow{1}{*}{RNN}  & 0.827 & N/A & StanfordNLP\_run5\\
\cline{2-3}
\hline
\multirow{1}{*}{Logical Inference}  & 0.827 & 32 & \cite{Johannes2014}\\
\cline{2-3}
\hline
\multirow{1}{*}{MaxEnt, SVM, kNN, GB, RF}  & 0.828 & 72 & \cite{Zhao2014ecnu}\\
\cline{2-3}
\hline
\multirow{1}{*}{WordEmbedding+MB-LSTM+Temp-CNN}  & 0.849 & 0 & \textbf{Our implementation}\\
\cline{2-3}
\hline
\multirow{1}{*}{CharCNN+MB-LSTM+Temp-CNN}  & 0.851 & 0 & \textbf{Our implementation}\\
\cline{2-3}
\hline
\end{tabular}
\end{center}
\caption{Semantic Relatedness Task Comparison.}
\label{tab:pear}
\end{table*}

\begin{table*}[!htb]
\begin{center}
\begin{tabular}{c|p{1cm}|p{1cm}|p{5cm}}
\hline
Method & \multicolumn{1}{c|}{Accuracy} & \multicolumn{1}{c|}{Features} & \multicolumn{1}{c}{Reported in}\\
\hline
\multirow{1}{*}{SVM}  & 0.823 & 41 & \cite{Thomas2014} \\
\cline{2-3}
\hline
\multirow{1}{*}{Decision tree}  & 0.831 & 214 & \cite{Jimenez2014}\\
\cline{2-3}
\hline
\multirow{1}{*}{MaxEnt, SVM, kNN, GB, RF}  & 0.836 & 72 & \cite{Zhao2014ecnu}\\
\cline{2-3}
\hline
\multirow{1}{*}{WordEmbedding+MB-LSTM+Temp-CNN}  & 0.838 & 0 & \textbf{Our implementation}\\
\cline{2-3}
\hline
\multirow{1}{*}{CharCNN+MB-LSTM+Temp-CNN}  & 0.842 & 0 & \textbf{Our implementation}\\
\cline{2-3}
\hline
\multirow{1}{*}{MaxEnt}  & 0.846 & 137 & \cite{Lai2014} \\
\cline{2-3}
\hline
\end{tabular}
\end{center}
\caption{Textual Entailment Task Comparison.}
\label{tab:acc}
\end{table*}

\subsection{Tree LSTM vs Sequence LSTM}
In this experiment, we will compare tree LSTM with sequential LSTM.  A limitation of the sequence LSTM architectures is that they only allow for strictly sequential information propagation. However, tree LSTMs allow richer network topologies where each LSTM unit is able to incorporate information from multiple child units.  As in standard LSTM units, each Tree-LSTM unit (indexed by $j$) contains input and output gates $i_j$ and $o_j$, a memory cell $c_j$ and hidden state $h_j$. The difference between the standard LSTM unit and tree LSTM units is that gating vectors and memory cell updates are dependent on the states of possibly many child units. Additionally, instead of a single forget gate, the tree LSTM unit contains one forget gate $f_{jk}$ for each child $k$. This allows the tree LSTM unit to selectively incorporate information from each child.

We use dependency tree child-sum tree LSTM proposed by~\cite{Tai2015} as our baseline. Given a tree, let $C(j)$ denote the set of children of node $j$. The child-sum tree LSTM transition equations are the following:
\begin{eqnarray}
\bold{\tilde{h}}_j &=& \sum_{k \in C(j) } h_k \,,\nonumber\\
\bold{i}_j &=& \sigma (\bold{W^{(i)}}\bold{x}_j + \bold{U^{(i)}}\bold{\tilde{h}}_{j} + \bold{b^{(i)}} ) \,,\nonumber\\
\bold{f}_{jk} &=& \sigma (\bold{W^{(f)}}\bold{x}_j + \bold{U^{(f)}}\bold{h}_{k} + \bold{b^{(f)}} ) \,,\nonumber\\
\bold{o}_j &=& \sigma (\bold{W^{(o)}}\bold{x}_j + \bold{U^{(o)}}\bold{\tilde{h}}_{j} + \bold{b^{(o)}} ) \,,\nonumber\\
\bold{u}_j &=& \tanh (\bold{W^{(u)}}\bold{x}_j + \bold{U^{(u)}}\bold{\tilde{h}}_{j} + \bold{b^{(u)}} ) \,,\nonumber\\
\bold{c}_j &=& \bold{i}_j \odot \bold{u}_j + \bold{f}_{jk} \odot \bold{c}_{k}\,,\nonumber\\
\bold{h}_j &=& \bold{o}_j \odot \tanh(\bold{c}_j) \,.
\end{eqnarray}

Table~\ref{tab:tg2} show the comparisons between tree and sequential based methods. We can see that, if we don't deploy CNN, simple Tree LSTM yields better result than traditional LSTM, but worse than Bidirectional LSTM. This is reasonable due to the fact that Bidirectional LSTM can enhance sentence representation by concatenating forward and backward representations. We found that adding CNN layer will decrease the accuracy in this scenario. Because when feeding into CNN, we have to reshape the feature planes otherwise convolution will not work. For example, we set convolution kernel width as 2, the input 2D tensor will have the shape lager than 2. To boost performance with CNN, we need more matching features. We found Multi-layer Bidirectional LSTM can incorporate more features and achieve best performance compared with single-layer Bidirectional LSTM.

\begin{table}[!htb]
\begin{center}
\begin{tabular}{c|p{1.5cm}|p{1.3cm}}
\hline
Method  & Accuracy & Pearson \\
\hline
Dep-Tree LSTM & 0.833 & 0.849 \\
\hline
Dep-Tree LSTM + CNN & 0.798 & 0.822\\
\hline
LSTM & 0.812 & 0.833  \\
\hline
LSTM + CNN & 0.776  & 0.810 \\
\hline
1-Bidirectional LSTM & 0.834 & 0.848  \\
\hline
1-Bidirectional LSTM+ CNN & 0.821 & 0.846\\
\hline
\end{tabular}
\end{center}
\caption{Results of Tree LSTM vs Sequence LSTM on auxiliary char embedding.}
\label{tab:tg2}
\end{table}

\section{Related Work}
Existing neural sentence models mainly fall into two groups: convolutional neural networks (CNNs) and recurrent neural networks (RNNs). In regular 1D CNNs~\cite{Collobert2011,kalchbrenner2013,Kim2014}, a fixed-size window slides over time (successive words in sequence) to extract local features of a sentence; then they pool these features to a vector, usually taking the maximum value in each dimension, for supervised learning. The convolutional unit, when combined with max-pooling, can act as the compositional operator with local selection mechanism as in the recursive autoencoder~\cite{Socher2011}. However, semantically related words that are not in one filter can't be captured effectively by this shallow architecture. \cite{Kalchbrenner2014} built deep convolutional models so that local features can mix at high-level layers. However, deep convolutional models may result in worse performance~\cite{Kim2014}. 

On the other hand, RNN can take advantage of the parsing or dependency tree of sentence structure information~\cite{Socher2011,Socher2014grounded}. \cite{Mohit2014} used dependency-tree recursive neural network to map text descriptions to quiz answers. Each node in the tree is represented as a vector; information is propagated recursively along the tree by some elaborate semantic composition. One major drawback of RNNs is the long propagation path of information near leaf nodes. As gradient may vanish when propagated through a deep path, such long dependency buries illuminating information under a complicated neural architecture, leading to the difficulty of training. To address this issue, \cite{Tai2015} proposed a Tree-Structured Long Short-Term Memory Networks. This motivates us to investigate multi-layer bidirectional LSTM that directly models sentence meanings without parsing for RTE task.

\section{Conclusions}
In this paper, we propose a new deep neural network architecture that jointly leverage pre-trained word embedding and character embedding to learn sentence meanings. Our new approach first generates two kinds of word sequence representations as inputs into bidirectional LSTM to learn sentence representation. After that, we construct matching features followed by another temporal CNN to learn high-level hidden matching feature representations. Our model shows that combining pre-trained word embeddings with auxiliary character-level embedding can improve the sentence representation. The enhanced sentence representation generated by multi-layer bidirectional LSTM will encapsulate the character and word levels informations. Furthermore, it may enhance matching features that generated by computing similarity measures on sentence pairs. Experimental results on benchmark datasets demonstrate that our new framework achieved the state-of-the-art performance compared with other deep neural networks based approaches.

\bibliographystyle{naaclhlt2016}
\bibliography{../../../BibTex/Bibliographies}

\begin{thebibliography}{}

\bibitem[\protect\citename{Bengio \bgroup et al.\egroup }1994]{Bengio1994}
Yoshua Bengio, Patrice Simard, and Paolo Fransconi.
\newblock 1994.
\newblock Learning long-term dependencies with gradient descent is difficult.
\newblock In {\em IEEE Transactions on Neural Networks 5(2)}.

\bibitem[\protect\citename{Bjerva \bgroup et al.\egroup }2014]{Johannes2014}
Johannes Bjerva, Johan Bos, Rob van~der Goot, and Malvina Nissim.
\newblock 2014.
\newblock {The Meaning Factory}: Formal semantics for recognizing textual
  entailment and determining semantic similarity.
\newblock In {\em Proceedings of SemEval 2014: International Workshop on
  Semantic Evaluation.}

\bibitem[\protect\citename{Collobert \bgroup et al.\egroup
  }2011]{Collobert2011}
Ronan Collobert, Jason Weston, L{\'e}on Bottou, Michael Karlen, Koray
  Kavukcuoglu, and Pavel Kuksa.
\newblock 2011.
\newblock Natural language processing (almost) from scratch.
\newblock {\em The Journal of Machine Learning Research}, 12:2493--2537.

\bibitem[\protect\citename{Duchi \bgroup et al.\egroup }2011]{Duchi2011}
John Duchi, Elad Hazan, and Yoram Singer.
\newblock 2011.
\newblock Adaptive subgradient methods for online learning and stochastic
  optimization.
\newblock {\em The Journal of Machine Learning Research}, 12:2121--2159.

\bibitem[\protect\citename{Graves \bgroup et al.\egroup }2013]{Alex2013}
Alex Graves, Navdeep Jaitly, and Abdel rahman Mohamed.
\newblock 2013.
\newblock Hybrid speech recognition with deep bidirectional lstm.
\newblock In {\em IEEE Workshop on Au- tomatic Speech Recognition and
  Understanding (ASRU)}, pages 273--278.

\bibitem[\protect\citename{Hochreiter and Schmidhuber}1998]{Sepp1998}
Sepp Hochreiter and J\"urgen Schmidhuber.
\newblock 1998.
\newblock Long short-term memory.
\newblock In {\em Neural Computation 9(8)}.

\bibitem[\protect\citename{Iyyer \bgroup et al.\egroup }2014]{Mohit2014}
Mohit Iyyer, Jordan Boyd-Graber, Leonardo Claudino, Richard Socher, and Hal
  {Daum\'e III}.
\newblock 2014.
\newblock A neural network for factoid question answering over paragraphs.
\newblock In {\em Empirical Methods in Natural Language Processing}.

\bibitem[\protect\citename{Jimenez \bgroup et al.\egroup }2014]{Jimenez2014}
Sergio Jimenez, George Duenas, Julia Baquero, and Alexander Gelbukh.
\newblock 2014.
\newblock {UNAL-NLP}: Combining soft cardinality features for semantic textual
  similarity, relatedness and entailment.
\newblock In {\em Proceedings of SemEval 2014: International Workshop on
  Semantic Evaluation.}

\bibitem[\protect\citename{Kalchbrenner and Blunsom}2013]{kalchbrenner2013}
Nal Kalchbrenner and Phil Blunsom.
\newblock 2013.
\newblock Recurrent continuous translation models.
\newblock In {\em Proceedings of the 2013 Conference on Empirical Methods in
  Natural Language Processing}, pages 1700--1709, Seattle, Washington, USA.
  Association for Computational Linguistics.

\bibitem[\protect\citename{Kalchbrenner \bgroup et al.\egroup
  }2014]{Kalchbrenner2014}
Nal Kalchbrenner, Edward Grefenstette, and Phil Blunsom.
\newblock 2014.
\newblock A convolutional neural network for modelling sentences.
\newblock {\em Proceedings of the 52nd Annual Meeting of the Association for
  Computational Linguistics}.

\bibitem[\protect\citename{Kim \bgroup et al.\egroup }2016]{kim2015character}
Yoon Kim, Yacine Jernite, David Sontag, and Alexander~M Rush.
\newblock 2016.
\newblock Character-aware neural language models.
\newblock In {\em Thirtieth AAAI Conference on Artificial Intelligence}.

\bibitem[\protect\citename{Kim}2014]{Kim2014}
Yoon Kim.
\newblock 2014.
\newblock Convolutional neural networks for sentence classification.
\newblock In {\em Proceedings of the 2014 Conference on Empirical Methods in
  Natural Language Processing (EMNLP)}, pages 1746--1751, Doha, Qatar.
  Association for Computational Linguistics.

\bibitem[\protect\citename{Lai and Hockenmaier}2014]{Lai2014}
Alice Lai and Julia Hockenmaier.
\newblock 2014.
\newblock Illinois-lh: A denotational and distributional approach to semantics.
\newblock In {\em Proceedings of SemEval 2014: International Workshop on
  Semantic Evaluation.}

\bibitem[\protect\citename{Mikolov \bgroup et al.\egroup }2013]{Mikolov2013}
Tomas Mikolov, Ilya Sutskever, Kai Chen, Greg~S Corrado, and Jeff Dean.
\newblock 2013.
\newblock Distributed representations of words and phrases and their
  compositionality.
\newblock In {\em Advances in Neural Information Processing Systems}, pages
  3111--3119.

\bibitem[\protect\citename{Proisl and Evert}2014]{Thomas2014}
Thomas Proisl and Stefan Evert.
\newblock 2014.
\newblock Robust semantic similarity at multiple levels using maximum weight
  matching.
\newblock In {\em Proceedings of SemEval 2014: International Workshop on
  Semantic Evaluation.}

\bibitem[\protect\citename{Socher \bgroup et al.\egroup
  }2011a]{Socher2011dynamic}
Richard Socher, Eric~H Huang, Jeffrey Pennin, Christopher~D Manning, and
  Andrew~Y Ng.
\newblock 2011a.
\newblock Dynamic pooling and unfolding recursive autoencoders for paraphrase
  detection.
\newblock In {\em Advances in Neural Information Processing Systems}, pages
  801--809.

\bibitem[\protect\citename{Socher \bgroup et al.\egroup }2011b]{Socher2011}
Richard Socher, Jeffrey Pennington, Eric~H Huang, Andrew~Y Ng, and
  Christopher~D Manning.
\newblock 2011b.
\newblock Semi-supervised recursive autoencoders for predicting sentiment
  distributions.
\newblock In {\em Proceedings of the Conference on Empirical Methods in Natural
  Language Processing}, pages 151--161. Association for Computational
  Linguistics.

\bibitem[\protect\citename{Socher \bgroup et al.\egroup
  }2014]{Socher2014grounded}
Richard Socher, Andrej Karpathy, Quoc~V Le, Christopher~D Manning, and Andrew~Y
  Ng.
\newblock 2014.
\newblock Grounded compositional semantics for finding and describing images
  with sentences.
\newblock {\em Transactions of the Association for Computational Linguistics}.

\bibitem[\protect\citename{Tai \bgroup et al.\egroup }2015]{Tai2015}
Kai~Sheng Tai, Richard Socher, and Christopher~D Manning.
\newblock 2015.
\newblock Improved semantic representations from tree-structured long
  short-term memory networks.
\newblock {\em arXiv preprint arXiv:1503.00075}.

\bibitem[\protect\citename{Yin and Schutze}2015]{Yin2015Multi}
Wenpeng Yin and Hinrich Schutze.
\newblock 2015.
\newblock Multigrancnn: An architecture for general matching of text chunks on
  multiple levels of granularity.
\newblock In {\em Proceedings of th 53rd Annual Meeting of the Association for
  Computational Linguistics}, pages 63--73.

\bibitem[\protect\citename{Zhang \bgroup et al.\egroup
  }2015]{zhang2015character}
Xiang Zhang, Junbo Zhao, and Yann LeCun.
\newblock 2015.
\newblock Character-level convolutional networks for text classification.
\newblock In {\em Advances in Neural Information Processing Systems}, pages
  649--657.

\bibitem[\protect\citename{Zhao \bgroup et al.\egroup }2014]{Zhao2014ecnu}
Jiang Zhao, Tian~Tian Zhu, and Man Lan.
\newblock 2014.
\newblock {ECNU}: One stone two birds: Ensemble of heterogenous measures for
  semantic relatedness and textual entailment.
\newblock In {\em Proceedings of SemEval 2014: International Workshop on
  Semantic Evaluation.}

\end{thebibliography}

\end{document}